\documentclass[11pt,a4paper]{article}
\usepackage[hyperref]{emnlp-ijcnlp-2019}
\usepackage{times}
\usepackage{latexsym}
\usepackage{amsmath}
\usepackage{amssymb}
\usepackage{amsfonts}
\usepackage[mathscr]{euscript}
\usepackage{graphicx}
\usepackage[skip=5pt]{caption}
\usepackage{subcaption}
\usepackage{bm}
\usepackage{url}
\usepackage{xspace}
\usepackage{natbib}
\usepackage{tabularx}
\usepackage{booktabs}
\usepackage{multirow}
\usepackage{algorithm}
\usepackage{algpseudocode}
\usepackage{enumitem}
\usepackage{paralist}
\usepackage{eqparbox}
\newdimen{\algindent}
\usepackage{tikz-dependency}
\usetikzlibrary{automata,decorations.markings,arrows,positioning,matrix,calc,patterns,angles,quotes,calc}

\algnewcommand\LeftComment[2]{%
	\hspace{#1\algindent}$\triangleright$ \eqparbox{COMMENT}{#2} \hfill %
}

\newcommand{\secref}[1]{\S\ref{sec:#1}}
\newcommand{\appref}[1]{Appendix~\ref{app:#1}}

\newcommand{\figref}[1]{Fig.~\ref{fig:#1}}
\newcommand{\tabref}[1]{Table~\ref{tab:#1}}

\definecolor{DarkGreen}{RGB}{0,111,0}
\definecolor{DarkBlue}{RGB}{0,0,111}
\definecolor{DarkRed}{RGB}{111,0,0}
\definecolor{DarkOrange}{RGB}{200,111,0}

\aclfinalcopy 
\def\aclpaperid{1934} 

\newcommand{\draftonly}[1]{#1}
\newcommand{\draftcomment}[3]{\draftonly{\textcolor{#2}{[#3 ({\bf #1})]}}}
\newcommand{\roy}[1]{\draftcomment{Roy}{red}{#1}}
\newcommand{\nascomment}[1]{\draftcomment{Noah}{olive}{#1}}
\newcommand{\hao}[1]{\draftcomment{Hao}{blue}{#1}}
\newcommand{\jdcomment}[1]{\draftcomment{Jesse}{teal}{#1}}

\newcommand{\com}[1]{} 
\newcommand{\lmcom}[1]{} 
\newcommand{\resolved}[1]{} 
\newcommand{\camready}[1]{$@@@$ #1} 
\newcommand{\interalia}[1]{\citep[\emph{inter alia}]{#1}}

\title{RNN Architecture Learning with Sparse Regularization}

\author{Jesse Dodge$^\clubsuit$ \quad 
Roy Schwartz$^\spadesuit$$^\diamondsuit$ \quad
Hao Peng$^\spadesuit$  \quad 
Noah A. Smith$^\spadesuit$$^\diamondsuit$ \\
  $^\clubsuit$Language Technologies Institute, Carnegie Mellon
  University, Pittsburgh, PA, USA\\ 
  $^\diamondsuit$Allen Institute for Artificial Intelligence, Seattle, WA, USA \\
  $^\spadesuit$Paul G. Allen School of Computer Science \&
  Engineering, University of Washington, Seattle, WA, USA\\
  {\tt jessed@cs.cmu.edu\quad roys@allenai.org \quad \{hapeng,nasmith\}@cs.washington.edu}
}

\date{}

\begin{document}
\maketitle
\com{\roy{authors}}
\resolved{\nascomment{title is unfriendly}}
\resolved{\nascomment{I like that the title is short and clean, but when I
  searched it on google scholar a bunch of things from 10 years ago
  came up.  I think we need to get RNN or neural or something like
  that into the title}}

\begin{abstract}
\resolved{\nascomment{I don't love the title; ``doing more with less'' just seems really vague and doesn't evoke anything in particular}}
\resolved{\hao{it seems that we focus on model compression now. do we want to make this point more straightforward? it seems to me that we go back and forth between sparsifying and compressing NLP models.}}
Neural models for NLP typically use large numbers of parameters to reach state-of-the-art performance, which can lead to excessive memory usage and increased runtime\com{ and large variance in performance}.
\resolved{\hao{can we find any pointer to support the large variance argument?
	actually, i know several work supporting the opposite,
	see https://arxiv.org/pdf/1712.09203.pdf and the cited works therein.}}
We present a structure learning method for learning sparse, parameter-efficient NLP models.
Our method applies group lasso to rational RNNs \cite{rational-recurrences}, a family of models that is closely connected to weighted finite-state automata (WFSAs).
We take advantage of rational RNNs' natural grouping of the weights, so the group lasso penalty directly removes WFSA states, substantially reducing the number of parameters in the model.
Our experiments on a number of sentiment analysis datasets\lmcom{ and one language modeling dataset}, using both GloVe and BERT embeddings, show that our approach learns neural structures which have fewer parameters without sacrificing performance relative to parameter-rich baselines.
Our method also highlights the interpretable properties of rational RNNs.
We show that sparsifying such models makes them easier to visualize, and we present models that rely exclusively on as few as three WFSAs after pruning more than 90\% of the weights.
 We publicly release our code.\footnote{\url{https://github.com/dodgejesse/sparsifying_regularizers_for_RRNNs}\resolved{\roy{link to code}.}}
\end{abstract}

\section{Introduction}

\resolved{\nascomment{it's worth stepping back and thinking what papers  our title/abstract will make readers think of.  then we should cite those in related work (or a more natural place).  one example that comes to mind is sparse attention / sparsemax}}

State-of-the-art neural models for NLP are heavily parameterized, requiring hundreds of millions \cite{Devlin:2018} and even billions \cite{Radford:2019} of parameters.
While over-parameterized models can sometimes be easier to train \cite{Livni:2014}, they may also introduce memory problems on small devices and lead to increased carbon emission \cite{Strubell:2019,Schwartz:2019}.

In feature-based NLP, structured-sparse regularization, in particular the group lasso \cite{group_lasso}, has been proposed as a method to reduce model size while preserving performance \cite{Martins:2011}.
But, in neural NLP, some of the most widely used models---LSTMs \cite{Hochreiter:1997} and GRUs \cite{Cho:2014}---do not have an obvious, intuitive notion of ``structure'' in their parameters (other than, perhaps, division into layers), so the use of structured sparsity at first may appear incongruous.

\com{
We propose a structure learning method that offers
parameter-efficient models\resolved{\nascomment{this is a bit misleading.  the models are sparse in a particular sense, but they are still densely connected NNs.  let's be more precise}} which perform similarly to parameter-rich alternatives. 
Our approach is based on structured-sparse regularization, in particular the group lasso \cite{group_lasso}, which has been successfully applied in feature-based NLP \cite{Martins:2011}, as well as in \camready{other subfields of AI such as }computer vision \cite{morphnet}.
The most widely used neural networks in NLP---LSTM \cite{Hochreiter:1997} and GRU \cite{Cho:2014}---do not have an obvious, intuitive notion of ``structure'' in the parameters (other than, perhaps, division into layers), so the use of structured sparsity at first may appear incongruous.
}

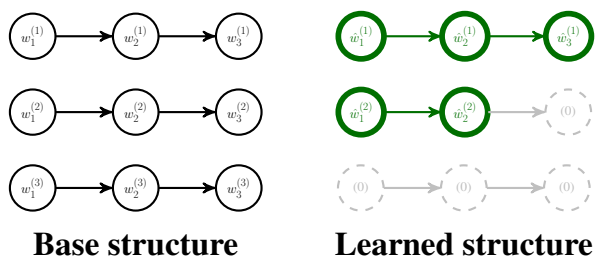
\begin{figure}[!t]
\vspace{-.3cm}

\begin{center}
\begin{tikzpicture}[remember picture,minimum size=3cm,font=\Huge,->,>=stealth',auto,node distance=1cm,
  thick,main node/.style={circle,draw,font=\sffamily\Large\bfseries}]
\newcommand{\ratio}[0]{0.2}
\newcommand{\dist}[0]{.75}
\newcommand{\distb}[0]{1}
\newcommand{\distc}[0]{.4}
\newcommand{\textsize}{\footnotesize}
\newcommand{\colora}[0]{black}
\newcommand{\colorb}[0]{DarkGreen}
\newcommand{\colorc}[0]{lightgray}

\node[circle,draw,\colora,scale=\ratio] (h11) at (-5,0) {$w^{(1)}_1$};
\node[circle,draw,\colora,scale=\ratio, right = \dist cm of h11] (h12)  {$w^{(1)}_2$};
\node[circle,draw,\colora,scale=\ratio, right = \dist cm of h12] (h13)  {$w^{(1)}_3$};
\node[circle,line width=.75mm,draw,\colorb,scale=\ratio, right = \distb cm of h13] (h14)  {$\hat{w}^{(1)}_1$};
\node[circle,line width=.75mm,draw,\colorb,scale=\ratio, right = \dist cm of h14] (h15)  {$\hat{w}^{(1)}_2$};
\node[circle,line width=.75mm,draw,\colorb,scale=\ratio, right = \dist cm of h15] (h16)  {$\hat{w}^{(1)}_3$};

\node[circle,draw,\colora,scale=\ratio, below = \distc of h11] (h21) {$w^{(2)}_1$};
\node[circle,draw,\colora,scale=\ratio, right = \dist cm of h21] (h22)  {$w^{(2)}_2$};
\node[circle,draw,\colora,scale=\ratio, right = \dist cm of h22] (h23)  {$w^{(2)}_3$};
\node[circle,line width=.75mm,draw,\colorb,scale=\ratio, right = \distb cm of h23] (h24)  {$\hat{w}^{(2)}_1$};
\node[circle,line width=.75mm,draw,\colorb,scale=\ratio, right = \dist cm of h24] (h25)  {$\hat{w}^{(2)}_2$};
\node[circle,draw,dashed,\colorc,scale=\ratio, right = \dist cm of h25] (h26)  {(0)};

\node[circle,draw,\colora,scale=\ratio, below = \distc of h21] (h31) {$w^{(3)}_1$};
\node[circle,draw,\colora,scale=\ratio, right = \dist cm of h31] (h32)  {$w^{(3)}_2$};
\node[circle,draw,\colora,scale=\ratio, right = \dist cm of h32] (h33)  {$w^{(3)}_3$};
\node[circle,draw,dashed,\colorc,scale=\ratio, right = \distb cm of h33] (h34)  {(0)};
\node[circle,draw,dashed,\colorc,scale=\ratio, right = \dist cm of h34] (h35)  {(0)};
\node[circle,draw,dashed,\colorc,scale=\ratio, right = \dist cm of h35] (h36)  {(0)};
\node[circle,line width=.75mm,scale=0.5, below = -1cm  of h32] (h37)  {{\fontfamily{arial}\selectfont {\bf Base structure}}};
\node[circle,line width=.75mm,scale=0.5, below = -1.35cm  of h35] (h38) {{\fontfamily{arial}\selectfont {\bf Learned structure}}};

  \path[\colora, every node/.style={font=\sffamily\textsize}]
     (h11) edge node [right] {} (h12)
     (h12) edge node [right] {}  (h13);

  \path[\colorb, every node/.style={font=\sffamily\textsize}]
     (h14) edge node [right] {} (h15)
     (h15) edge node [right] {}  (h16);

  \path[\colora, every node/.style={font=\sffamily\textsize}]
     (h21) edge node [right] {} (h22)
     (h22) edge node [right] {}  (h23);

  \path[\colorb, every node/.style={font=\sffamily\textsize}]
     (h24) edge node [right] {} (h25);

  \path[\colorc, every node/.style={font=\sffamily\textsize}]
     (h25) edge node [right] {} (h26);

 \path[every node/.style={font=\sffamily\textsize}]
     (h31) edge node [right] {} (h32)
     (h32) edge node [right] {}  (h33);

  \path[\colorc, every node/.style={font=\sffamily\textsize}]
     (h34) edge node [right] {} (h35)
     (h35) edge node [right] {} (h36);


\end{tikzpicture}
\end{center}

\setlength{\abovecaptionskip}{-.5cm}
\vspace{-1cm}
\caption{\label{fig:our-method}
Our approach learns a sparse structure (right hand side) of a base rational RNN (left hand side) where each hidden unit corresponds to a WFSA (in this example, three hidden units, represented by the three rows). 
Grayed-out, dashed states are removed from the model, while retained states are marked in {\textcolor{DarkGreen}{\bf bold green}}.
}
\end{figure}

In this paper we show that group lasso can be successfully applied to neural NLP models. 
We focus on a family of neural models 
for which the hidden state exhibits a natural structure: rational RNNs \cite{rational-recurrences}.
In a rational RNN, the value of each hidden dimension is the score of a weighted finite-state automaton (WFSA) on (a prefix of) the input vector sequence.
This property offers a natural grouping of the transition function parameters for each WFSA.  
As shown by \citet{Schwartz:2018} and \citet{rational-recurrences}, a variety of state-of-the-art neural architectures  are rational \interalia{Lei:2017,Bradbury:2017,Foerster:2017},
so learning parameter-efficient rational RNNs is of practical value.
We also take advantage of the natural interpretation of rational RNNs as ``soft'' patterns \cite{Schwartz:2018}.


 \com{Unlike traditional WFSA-based models, rational recurrences compute their WFSA scores based on word vectors (rather than word symbols), and can thus be seen as capturing ``soft'' patterns \cite{Schwartz:2018}.}
\resolved{\hao{the above statement is not true. so i changed it to the following:}}
\resolved{\nascomment{maybe add a sentence saying why this is appealing, and also clarify that these WFSAs match word vectors, not symbols, so they are ``soft'' and cite SoPa}}

\resolved{\nascomment{suggest reworking this to motivate the particular kind of sparsity we seek:  eliminate states/transitions we do not need}}

We apply a group lasso penalty to the WFSA parameters of  rational RNNs during training, where each group is comprised of the parameters associated with one state in one WFSA (\figref{our-method}; \secref{models}). \resolved{\hao{apply ... penalty to the parameters? instead of loss}}This penalty pushes the parameters in some groups to zero, effectively eliminating \resolved{\nascomment{you will not need to talk about dropout here if you use a different verb, how about ``eliminate''? :-)  now the footnote is unnecessary } }them, and making the WFSA smaller.\com{\footnote{Our approach should not be confused with dropout \cite{}. In dropout, individual random parameters are being dropped during training. Our approach exploits the structure of rational RNNs and drops groups of parameters, both at train and test time.}}
\resolved{\hao{a very minor point: do we want to use a footnote to clarify how this is different from dropout?}}When all of the states for a given WFSA are eliminated, the WFSA is removed entirely, so this approach can be viewed as learning the number of WFSAs (i.e., the RNN hidden dimension) as well as their size. 
We then retain \resolved{\hao{retrain?} }the sparse structure, which results in a much smaller model in terms of parameters. 
\resolved{\hao{length->structure.}}

We experiment with four text classification benchmarks~(\secref{experiments}), 
using both GloVe and BERT embeddings.\lmcom{, as well as one language modeling benchmark}
As we increase the regularization strength, we end up with smaller models.
These models have a better tradeoff between the number of parameters and model performance
compared to setting the number of WFSAs and their lengths by hand or using hyperparameter search. 
In almost all cases, our approach results in models with fewer parameters \resolved{\nascomment{this seems like a weaker version of what you said above -- that we always get smaller models!}}and similar or better performance compared to our baselines.

In contrast to neural architecture search~\citep{Jozefowicz:2015,NAS}, \resolved{\nascomment{maybe also cite this one:}}
which can take several GPU years to learn an appropriate neural architecture, 
our approach requires only two training runs: 
one to learn the structure,
and the other to estimate its parameters. 
Other approaches \resolved{\nascomment{how can a structure learning approach ignore structure?  can we make this statement more precise?}} either ignore the structure of the model and only look at the value of individual weights \cite{DARTS,brain_damage, SNIP,Frankle:2019} or only use high-level structures like the number of layers of the network \cite{wen2016learning,group_lasso_nn,morphnet}.

Finally, our approach touches on another appealing property of rational RNNs---their interpretability. 
Each WFSA 
captures a ``soft" version of patterns like 
``such a great X'', and can be visualized as such \cite{Schwartz:2018}. 
By retaining a small number of WFSAs, model structures learned using our method can be visualized succinctly.
In~\secref{visualization} we show that some of our sentiment analysis models rely exclusively on as few as \emph{three} WFSAs.\footnote{That is, a rational RNN with hidden size 3. \resolved{\nascomment{is that correct?  aren't the hidden dimensions the states in the WFSAs?}}}\resolved{ with 8 main-path transitions \hao{``main-path'' never defined} in total. } 
\section{Method}\label{sec:models}

We describe the proposed method.
At a high level, we follow the standard practice for using $\ell_1$ regularization for sparsification \cite{wen2016learning}:

\begin{compactitem}
	\item[1.] Fit a model\com{ (\secref{rrnn})}
	on the training data, with the group lasso regularizer added
        to the loss during training (the parameters associated with
        one state comprise one group)\com{; see Eq.~\ref{eq:grouplasso}}.
	\item[2.] After convergence, eliminate the states whose
          parameters are zero.
	\item[3.] Finetune the resulting, smaller model, by minimizing
          the unregularized loss with respect to its parameters.
\end{compactitem}

\begin{figure}
	\centering
	\includegraphics[clip,trim=0cm 10cm 0cm 10.5cm, width=.98\columnwidth]{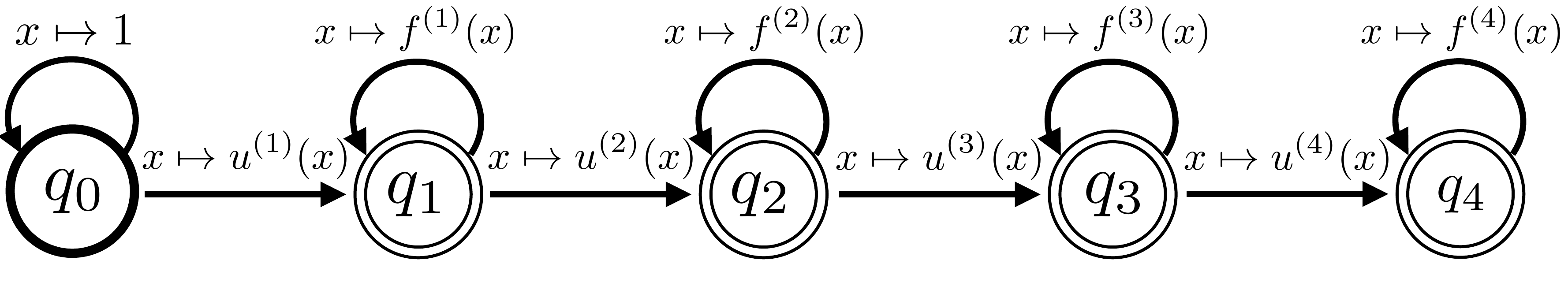}
	\caption{A 4-gram WFSA, from which we derive the rational RNN (\secref{models}). 
		The rational RNN's hidden states correspond to a set of WFSAs,
		each separately parameterized.
		We apply group lasso to each WFSA.}
	\label{fig:4gram_wfsa}
\end{figure}
In this work, we assume a single layer rational RNN, but our approach is equally applicable to multi-layer models.
For clarity of the discussion, we start with a one-dimensional
rational RNN (i.e., one based on a single WFSA only).
We then generalize to the $d$-dimensional case (computing the scores of $d$ WFSAs in parallel).

\paragraph{Rational recurrent networks}\label{sec:rrnn}
\resolved{\hao{TODO: something to justify 4-gram}}
Following \citet{rational-recurrences},
we parameterize the transition functions of WFSAs with neural networks, such that each transition (main path or self loop) defines a weighted function over the input word vector.
We consider a 5-state WFSA, diagrammed in~\figref{4gram_wfsa}.
\resolved{\hao{minor change here. check.}}

A path starts at $q_0$; 
at least four tokens must be consumed to reach $q_4$,
and in this sense it captures 4-gram ``soft'' patterns~\citep{rational-recurrences,Schwartz:2018}.
In addition to $q_4$, we also designate $q_1$, $q_2$, and $q_3$ as final states,
allowing for the interpolation between patterns of different lengths.\footnote{We found this to be more stable than using only $q_4$.}
The self-loop transitions over $q_1$, $q_2$, $q_3$, and $q_4$
aim to allow, but downweight, nonconsecutive patterns,
as the self-loop transition functions yield values between 0 and 1 (using a sigmoid function). 
The recurrent function is equivalent to applying the Forward dynamic programming algorithm~\citep{baum_statistical_1966}.
\com{To make the discussion self-contained, we concretely walk through the derivation below.
Given input string $\mathbf{x}=x_1\dots x_n$,
let $c_{t}^{(i)}$ (for any $t \le n$) denote the sum of scores of all paths ending in state $q_i$
after consuming prefix $x_1 \dots x_t$.
Denoting $u^{(i)}(x_t)$ by $u^{(i)}_t$, and $f^{(i)}(x_t)$ by $f^{(i)}_t$,
we have
\begin{subequations}\label{eq:4gram_recurrence}
	\begin{align}
	c_{t}^{(0)} &=1 \\
	c_{t}^{(i)} &= c_{t-1}^{(i)} \cdot f^{(i)}_t 
	+ c_{t-1}^{(i-1)}\cdot u^{(i)}_t,
	\end{align}
\end{subequations}
for $i \in \{1,2,3,4\}$.

The functions $f^{(i)}$ (representing self-loops) and $u^{(i)}$ (representing transitions) can be parameterized with neural networks.
Letting $\mathbf{z}_t$ denote the embedding vector for token $x_t$,
\begin{subequations}\label{eq:4gram_weights}
	\begin{align}
	f^{(i)}_t&=\sigma\bigl(\mathbf{w}^{(i)\top}\mathbf{z}_{t}\bigr),\\
	u^{(i)}_t&=(1-f^{(i)}_t) \cdot\mathbf{v}^{(i)\top} \mathbf{z}_{t},
	\end{align}
\end{subequations}
where $\mathbf{w}^{(i)}$ and $\mathbf{v}^{(i)}$ vectors are learned parameters.
In the interest of notational clarity, we suppress the bias terms in the affine transformations.

$c_t$, the total score of the prefix string $x_1 \dots x_t$, is calculated by summing
the scores of the paths ending in each of the final states:
\begin{align}
c_{t} = \sum_{i=1}^{4}c_{t}^{(i)}.
\end{align}
For a document of length $n$, $c_n$  is then used in downstream computation, e.g.,
fed into an MLP classifier.}

\paragraph{Promoting sparsity with group lasso}
\com{We now apply group lasso to promote
sparsity in a rational RNN, continuing with the same running example.}
We aim to learn a sparse rational model with fewer WFSA states.
This can be achieved by penalizing the parameters associated with a
given state, specifically the parameters associated with 
\emph{entering} that state, by a transition from another state
or a self-loop on that state.
For example, the parameters of the WFSA in \figref{4gram_wfsa} (excluding the word 
embedding parameters) 
are assigned to four nonoverlapping groups, one for each non-starting state.

During training\com{ to solve Eq.~\ref{eq:grouplasso}},
the regularization term will push all parameters toward zero, and some will converge
close to zero.\footnote{There are optimization methods that
  achieve ``strong'' sparsity 
\cite{parikhboyd2013}, where some parameters are exactly set to
  zero during training. Recent work has shown these approaches can converge in nonconvex settings \cite{reddiProximal}, but our experiments found them to be unstable.}
After convergence, we remove groups for which the $\ell_2$ norm falls below $\epsilon$.\footnote{We use 0.1. This threshold was
  lightly tuned in preliminary experiments on the validation set \resolved{\hao{on dev. set?} }and found to reliably remove those parameters which converged around zero without removing others.}\resolved{\roy{Idea we should try: use the lottery ticket strategy rather than simply the small values}}
The resulting smaller model is then finetuned by continuing training
without regularizing\com{, i.e., setting $\lambda = 0$}.
With our linear-structured WFSA, zeroing out the group
associated with a state in the middle effectively makes later states
inaccessible. 
While our approach offers no guarantee to remove states from the end
first (thus leaving no unreachable states),  in our experiments it
always did so.

\paragraph{$d$-dimensional Case}
\com{So far we discussed a one-dimensional model, i.e., a
rational RNN with one WFSA.} \resolved{\nascomment{again, I thought the
dimensionality was the total number of states, not the total number of WFSAs!}}
To construct a rational RNN with $d$ WFSAs (a $d$-dimensional model), we stack $d$ one-dimensional models, each of them separately parameterized.\com{and recovers a single dimension of the $d$ recurrent computation.}\com{\footnote{Such elementwise recurrent computation is \emph{not}
	a necessary condition for the model to be rational.
	We refer the readers to Section 4.3 of~\citet{rational-recurrences} for related discussion.
}}\com{Elementwise recurrent updates lead to desirable properties when
trained with group lasso.} The parameters of a $d$-dimensional rational model derived from the WFSA in \figref{4gram_wfsa} are organized into $4d$ groups, four for each dimension.
Since there is no direct interaction between different dimensions (e.g., through an affine transformation),
group lasso sparsifies each dimension/WFSA independently.
Hence the resulting rational RNN can consist of WFSAs of different sizes,
the number of which could be smaller than $d$ if any of the WFSAs have all states eliminated.
\resolved{\nascomment{I don't understand this, probaby cut:  This naturally connects to the interpolation of features of different orders,
		e.g., in a $n$-gram language model~\citep{some nlp text books?}.}}

One can\com{, of course} treat the numbers and sizes of WFSAs as hyperparameters~\cite{oncina1993learning,ron1994learning,higuera2010grammatical,Schwartz:2018}.
By eliminating states from WFSAs with group lasso,
we learn the WFSA structure \emph{while} fitting the models' parameters,
reducing the number of training cycles by reducing the number of tunable hyperparameters. \resolved{\hao{is this argument true? or useful at all?}
\hao{i'm trying to the below comment by this paragraph. 
	i'm not sure this is well done. 
	do we have better ways?}
\nascomment{I think here we should reiterate that this could also be
	accomplished by treating the number/size of WFSAs as a bunch of
	hyperparameters, and also referring to old papers about FSA
	structure learning.  then say explicitly that we don't want to run
	parameter estimation more times than necessary, and our method lets
	us learn the structure while we learn the parameters)}}

\section{Experiments}\label{sec:experiments}

We run sentiment analysis experiments\lmcom{two tasks: text classification and language modeling}.
We\lmcom{ follow the same procedure for both tasks: we} train the rational RNN models (\secref{models}) with group lasso regularization,
using increasingly large regularization strengths, resulting in increasingly compact models.
As the goal of our experiments is to demonstrate the ability of our approach to reduce the number of parameters, we only consider rational baselines:
the same rational RNNs trained without group lasso.\footnote{Rational RNNs have shown strong performance on the dataset we experiment with: a 2-layer rational model with between 100--300 hidden units obtained 92.7\% classification accuracy, substantially outperforming an LSTM baseline \cite{rational-recurrences}. The results of our models, which are single-layered and capped at 24 hidden units, are not directly comparable to these baselines, but are still within two points of the best result from that paper. }\resolved{\roy{dropped the footnote here. one reviewer complained about this point. it indeed sounds bad}}
\resolved{Instead, \hao{why `instead`?} }We manually tune the number and sizes of the baselines WFSAs, and\resolved{  \nascomment{say explicitly where the baseline stands with respect to the state of the art. if it's not the best published model, say how close it is.  head off any ``weak baseline'' complaints here}}
then compare the tradeoff curve between model size and accuracy.
We describe our experiments below. For more details, see \appref{experimental_details}.

\lmcom{\subsection{Text Classification}
We experiment with 4 sentiment analysis dataset. We train models both using type level embeddings and contextual, BERT embeddings.
}
\paragraph{Data} 
We experiment with the Amazon reviews binary sentiment classification dataset \cite{Blitzer:2007},\com{\footnote{\url{https://www.cs.jhu.edu/~mdredze/datasets/sentiment/}}} composed of 22 product categories. 
We examine the standard dataset ({\bf original\_mix}) comprised of a mixture of data from the different categories \cite{johnson2015effective}.\footnote{\url{http://riejohnson.com/cnn_data.html}}
We also examine three of the largest individual categories as separate datasets ({\bf kitchen}, {\bf dvd}, and {\bf books}), following \citet{johnson2015effective}.
The three category datasets do \emph{not} overlap with each other (though they do with {\bf original\_mix}), 
and are significantly different in size (see \appref{experimental_details}), so we can see how our approach behaves with different amounts of training data.
\resolved{\nascomment{it sounds like this is not the conventional way people do experiments on these datasets.  if that's the case, say so and justify it. otherwise reviewers will think we're downplaying that detail and possibly weaknesses of our approach.}}
\resolved{\hao{mention they are all binary classification?}}

\paragraph{Implementation details}
To classify text, we concatenate the scores computed by each WFSA, 
then feed this $d$-dimensional vector of scores into a linear binary classifier.
We use log loss.
We experiment with both 
type-level word embeddings (GloVe.6B.300d;~\citealp{Pennington:2014}) 
and contextual embeddings (BERT large;~\citealp{Devlin:2018}).\footnote{\url{https://github.com/huggingface/pytorch-pretrained-BERT}}\resolved{\hao{are these fine-tuned or fixed?}}
In both cases, we keep the embeddings fixed, so the vast majority of the learnable parameters are in the WFSAs. \resolved{\nascomment{I think you need to say that embeddings are never finetuned}}
We train models using GloVe embeddings on all datasets. Due to memory constraints we evaluate BERT embeddings (frozen, not fine-tuned) only on the smallest dataset ({\bf kitchen}).
\resolved{ \hao {`GloVe models' means models using GloVe, right?
 	also, `evaluate BERT' seems ambiguous: do we actually train the model using BERT as word embeddings?
 	}}
\paragraph{Baselines}
As baselines, we train five versions of each rational architecture without group lasso, using the same number of WFSAs as our regularized models (24 for GloVe, 12 for BERT).
Four of the baselines each use the same number of transitions for all WFSAs (1, 2, 3, and 4, corresponding to 2--5 states, and to 24, 48, 72, and 96 total transitions).
The fifth baseline has an equal mix of all lengths (6 WFSAs of each size for GloVe, leading to 60 total transitions, and 3 WFSAs of each size for BERT, leading to 30 total transitions).

Each transition in our model is independently parameterized, so the total number of transitions linearly controls the number of learnable parameters (in addition to the parameters in the embedding layer).



\definecolor{our_orange}{HTML}{FF7F0E}
\definecolor{our_blue}{HTML}{1F77B4}

\begin{figure*}[!t]
\includegraphics[clip,trim={0cm 0.45cm 0cm .35cm},scale=0.4]{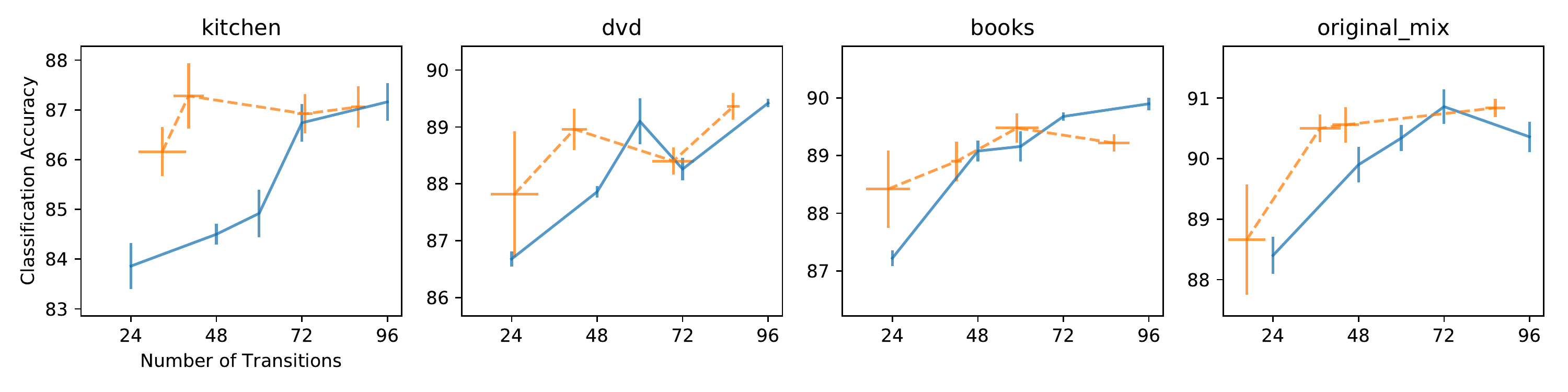}
\includegraphics[clip,trim={1.2cm 0.45cm 20cm 0.35cm},scale=0.4]{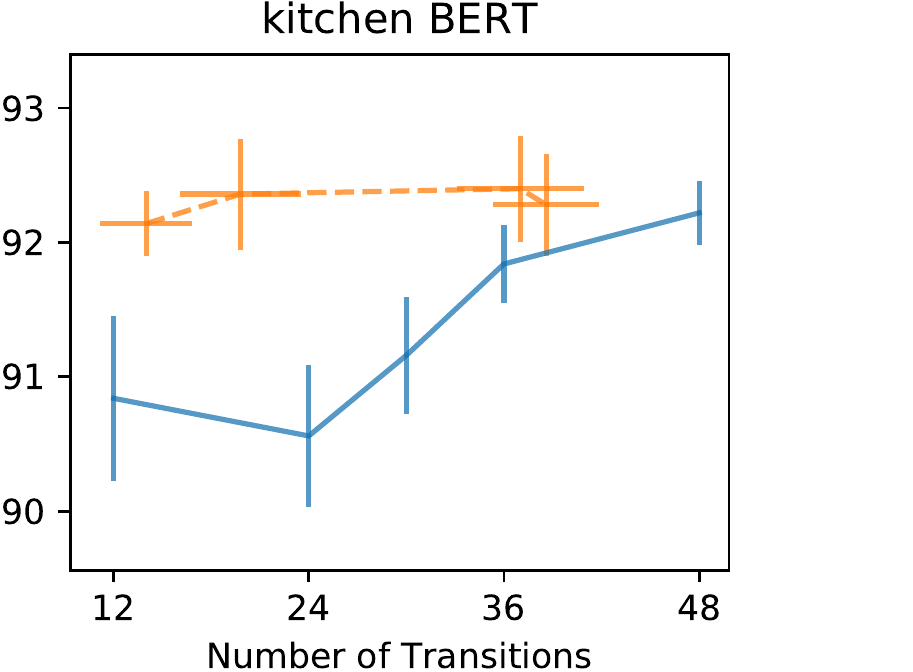}
\caption{\label{fig:textcat_results} Text classification with GloVe embeddings (four leftmost graphs) and BERT (rightmost):  accuracy ($y$-axis) vs.~number of WFSA transitions ($x$-axis). {\bf Higher} and to the {\bf left} is better. 
Our method ({\color{our_orange}{dashed orange line}}, varying regularization strength\resolved{\nascomment{check!}}) provides a better tradeoff than the baseline ({\textcolor{our_blue}{solid blue line}}\resolved{\roy{Jesse, can you please change the font color to match the color in the graph?}}, directly varying the number of transitions).
Vertical lines encode one standard deviation for accuracy, while horizontal lines encode one standard deviation in the number of transitions (applicable only to our method).
}
\end{figure*}

\com{
\begin{figure}
\centering
\includegraphics[clip,trim={0 0 20cm .75cm},scale=0.5]{fig/classification_accuracy_BERT.pdf}
\caption{\label{fig:textcat_results_bert} Text classification results using BERT embeddings on the {\bf kitchen} dataset.
}
\end{figure}
}

\paragraph{Results}

\resolved{\hao{can we say something like, \#parameters (excluding embeddings) linearly depend on \#transitions?
	can we move the thing in the `discussion` paragraph upfront, to justify why we plot \#transitions?}}
\figref{textcat_results} shows our classification test accuracy as a function of the total number of WFSA transitions in the model. 
We first notice that, as expected, the performance of our unregularized baselines improves as models are trained with more transitions (i.e., more parameters).

Compared to the baselines, training with group lasso provides a better tradeoff between performance and number of transitions.
In particular, our heavily regularized models perform substantially better than the unigram baselines, gaining between 1--2\% absolute improvements \resolved{\nascomment{what is the ``small'' baseline?} }in four out of five cases.
As our regularization strength decreases, we naturally gain less compared to our baselines, \resolved{\nascomment{rest of this sentence doesn't align with what I see.  maybe better to say ``never worse''?} }although still similar or better than the best baselines in four out of five cases.

\section{Visualization}\label{sec:visualization}

Using our method with a high regularization strength, 
the resulting sparse structures often contain only a handful of WFSAs.
In such cases, building on the interpretability of individual WFSAs,
we are able visualize every hidden unit, i.e., the entire model. 
To visualize a single WFSA $\mathscr{B}$, we follow \citet{Schwartz:2018} and compute the score of $\mathscr{B}$ on every phrase in the training corpus.
We then select the top and bottom scoring phrases for $\mathscr{B}$,\footnote{As each WFSA score is used as a feature that is 
fed to a linear classifier, 
negative scores are also meaningful.} and get a prototype-like description of the pattern representing $\mathscr{B}$.\footnote{
While the WFSA scores are the sum of all paths deriving a document (plus-times semiring), here we search for the max (or min) scoring one.
Despite the mismatch, a WFSA scores every possible path, and thus the max/min scoring path selection is still valid. 
As our examples show, many of these extracted paths are meaningful.
}
\com{which visualizes individual WFSAs (i.e., an RNN hidden unit).}\com{yet rational models typically consist of dozens of WFSAs,
making it indigestible for practitioners to visualize all WFSAs. 
Our approach helps to alleviate this problem. }

\newcommand{\highest}[0]{Top}
\newcommand{\bottom}[0]{Bottom}
\newcommand{\prefix}[0]{Patt.~}

\begin{table}[!t]
\small
\newcommand{\sloop}[1]{{\color{blue}{#1$_{\textit{SL}}$}}}
\setlength{\tabcolsep}{5.5pt}
\scriptsize
\begin{tabularx}{\linewidth}{c|c|lll}
  \toprule
\multicolumn{2}{c}{} &  {\bf transition$_1$} &     {\bf transition$_2$}        &       {\bf transition$_3$}  \\
  \toprule
  \multirow{10}{*}{\vspace{0.5cm} \prefix 1} &\multirow{5}{*}{\vspace{0.25cm} \highest} & not &     worth        &       \sloop{the time}       $<$/s$>$  \\
&  & not &     worth        &       \sloop{the 30}       $<$/s$>$  \\
&  & not &     worth        &       \sloop{it}       $<$/s$>$  \\
&  & not &     worth        &       \sloop{it}       $<$/s$>$  \\
\cline{2-5}
 &\multirow{5}{*}{\vspace{0.25cm} \bottom}  & extremely &     pleased        &       \sloop{\dots}       $<$/s$>$  \\
&  & highly &      pleased        &       \sloop{\dots}       $<$/s$>$  \\
&  & extremely &     pleased        &       \sloop{\dots}       $<$/s$>$  \\
&  & extremely &     pleased        &       \sloop{\dots}       $<$/s$>$  \\

  \midrule

  \multirow{10}{*}{\vspace{0.5cm} \prefix 2} &\multirow{5}{*}{\vspace{0.25cm} \highest} & miserable&     $<$/s$>$    &          \\ 
&& miserable      & \sloop{\dots}  $<$/s$>$    &          \\ 
 && miserable      & \sloop{\dots}  $<$/s$>$    &          \\ 
 && returned   & $<$/s$>$ &              \\ 
\cline{2-5}
 &\multirow{5}{*}{\vspace{0.25cm} \bottom} & superb&     $<$/s$>$    &          \\ 
 && superb&     $<$/s$>$    &          \\ 
 && superb&     $<$/s$>$    &          \\ 
 && superb&   \sloop{choice}   $<$/s$>$    &          \\ 

  \midrule

  \multirow{10}{*}{\vspace{0.5cm} \prefix 3} &\multirow{5}{*}{\vspace{0.25cm} \highest} & bad & \sloop{\dots} ltd & \sloop{\dots} buyer \\
 & & bad & \sloop{\dots} ltd & \sloop{\dots} buyer \\
 & & horrible & \sloop{\dots} hl4040cn & \sloop{\dots} expensive \\
 & & left & \sloop{\dots} ltd & \sloop{\dots} lens \\
 \cline{2-5}
 &\multirow{5}{*}{\vspace{0.25cm} \bottom}  & favorite & \sloop{\dots} ltd & \sloop{\dots} lens \\
 & & really & \sloop{\dots} ltd & \sloop{\dots} buyer \\
 & & really & \sloop{\dots} ltd & \sloop{\dots} buyer \\
 & & best & \sloop{\dots} hl4040cn & \sloop{\dots} expensive \\

  \bottomrule
\end{tabularx}
\caption{\label{tab:viza}
Visualization of a sparse rational RNN trained on {\bf original\_mix} containing only 3 WFSAs.
For each WFSA (i.e., pattern), we show the 4 top and bottom scoring phrases in the training corpus with this WFSA.
Each column represents one main-path transition, plus potential self-loops preceding it (marked like \sloop{this}). 
``\sloop{\dots}'' marks more than 2 self loops. ``$<$/s$>$'' marks an end-of-document token.
}
\end{table}

\resolved{
\nascomment{what is ``hl4040cn''? (looks like it's a printer model?)  is that actually an UNK in our models?}
\jdcomment{looking at the training data: it actually is a printer model. it only appears in one example. similarly, ``ltd'' appears only after the name of a company (in two examples). I'm not sure what to make of this, other than learning is a noisy process?}
}

\tabref{viza} visualizes a sparse rational RNN trained on {\bf original\_mix} with only \emph{three} WFSAs, 
(8 main-path transitions in total).\footnote{The test performance of this model is 88\%, 0.6\% absolute below the average of the five models reported in \figref{textcat_results}.}
The table shows that looking at the top scores of each WFSA, 
two of the patterns respectively capture the phrases ``\emph{not worth X $<$/s$>$}'' and ``\emph{miserable/returned X $<$/s$>$'}'.
Pattern 3 \resolved{\nascomment{(pattern 2) ?} }is not as coherent, but most examples do contain sentiment-bearing words such as \emph{bad, horrible,} or \emph{best}. 
This might be the result of the tuning process of the sparse rational structure simply learning a collection of words, rather than coherant phrases.
As a result, this WFSA is treated as a unigram pattern rather than
a trigram.
The lowest scoring phrases show a similar trend.
\com{\roy{Looking at the third pattern again, it does actually seem coherent. maybe the problem is that it doesn't look like a pattern but a collection of words? Jesse, can you please rephrase?}}
\appref{visualization} shows the same visualization for another sparse rational RNN containing only four WFSAs and 11 main-path transitions, trained with BERT embeddings.

We observe another interesting trend:
two of the three patterns
prefer expressions that appear near the end of the document. 
This could result from the nature of the datasets (e.g., many reviews end
with a summary, containing important sentiment information), 
and/or our rational models' recency preference.
More specifically,
the first self loop has weight $1$, and hence the model is not
penalized for taking self loops before the match;
in contrast, the
weights of the last self loop
take values in $(0,1)$ due to the sigmoid,
forcing a penalty for earlier phrase matches.\footnote{Changing this behavior could be easily done by fixing the final self-loop to $1$ as well.}


\com{We also observe that the patterns seem to be targeting fixed (rather
than soft) $n$-grams. For instance, the top and bottom four matches of
patterns 1 and 3 all match the same expressions. \nascomment{I don't
  buy it.  you're only looking at 10 examples!}
Finally, comparing the top scores vs.~the bottom scores, we see that
each WFSA is learning (at least) two different patterns:
one for phrases with high scores, and the other for ones
with low (negative) scores. 
Still, given the small
number of WFSAs used by the model, we are able to visualize
all learned (soft) pattern in a single table. 
}


\section{Conclusion}

We presented a method for learning parameter-efficient RNNs.
Our method applies group lasso regularization on rational RNNs, which are strongly connected to weighted finite-state automata, and thus amenable to learning with structured sparsity.
Our experiments on four text classification datasets, using both GloVe and BERT embeddings, show that our sparse models provide a better performance/model size tradeoff.
We hope our method will facilitate the development of ``thin'' NLP models, that are faster, consume less memory, and are interpretable \cite{Schwartz:2019}.\resolved{\nascomment{and interpretable?}}

\resolved{
\jdcomment{things i would say as a reviewer: so you have a big matrix that is (num wfsa) X (num ngrams). your baseline is reducing the (num ngrams), but you didn't try reducing (num wfsa). your claim is that reducing the size of that matrix by reducing both the number of WFSAs and the number of ngrams (using the regularizer to balance these reducitions) is the right approach, but you didn't try any baselines doing this (assuming we leave out the even-mix baseline because it was too strong).}

\jdcomment{why not compare against all the other approaches which can be used to reduce the size of a neural network?}}

\section*{Acknowledgments}
This work was completed while the first author was an intern at the Allen Institute for Artificial Intelligence.  
The authors thank Pradeep Dasigi, Gabriel Stanovsky, and Elad Eban for their discussion.
In addition, the authors thank the members of the Noah’s ARK group at the University of Washington, the researchers at the Allen Institute for Artificial Intelligence, and the anonymous reviewers for their valuable feedback.
This work was supported in part by a hardware gift from NVIDIA Corporation.

\bibliography{acl2019}

\begin{thebibliography}{34}
\expandafter\ifx\csname natexlab\endcsname\relax\def\natexlab#1{#1}\fi

\bibitem[{Bahdanau et~al.(2015)Bahdanau, Cho, and Bengio}]{Bahdanau:2015}
Dzmitry Bahdanau, Kyunghyun Cho, and Yoshua Bengio. 2015.
\newblock Neural machine translation by jointly learning to align and
  translate.
\newblock In \emph{Proc. of ICLR}.

\bibitem[{Baum and Petrie(1966)}]{baum_statistical_1966}
Leonard~E. Baum and Ted Petrie. 1966.
\newblock \href {https://doi.org/10.1214/aoms/1177699147} {Statistical
  inference for probabilistic functions of finite state {Markov} chains}.
\newblock \emph{The Annals of Mathematical Statistics}, 37(6).

\bibitem[{Blitzer et~al.(2007)Blitzer, Dredze, and Pereira}]{Blitzer:2007}
John Blitzer, Mark Dredze, and Fernando Pereira. 2007.
\newblock \href {http://aclweb.org/anthology/P07-1056} {Biographies,
  {Bollywood}, boom-boxes and blenders: Domain adaptation for sentiment
  classification}.
\newblock In \emph{Proc. of ACL}.

\bibitem[{Bradbury et~al.(2017)Bradbury, Merity, Xiong, and
  Socher}]{Bradbury:2017}
James Bradbury, Stephen Merity, Caiming Xiong, and Richard Socher. 2017.
\newblock Quasi-recurrent neural network.
\newblock In \emph{Proc. of ICLR}.

\bibitem[{Cho et~al.(2014)Cho, Van~Merri{\"e}nboer, Gulcehre, Bahdanau,
  Bougares, Schwenk, and Bengio}]{Cho:2014}
Kyunghyun Cho, Bart Van~Merri{\"e}nboer, Caglar Gulcehre, Dzmitry Bahdanau,
  Fethi Bougares, Holger Schwenk, and Yoshua Bengio. 2014.
\newblock \href {http://www.aclweb.org/anthology/D14-1179} {Learning phrase
  representations using {RNN} encoder-decoder for statistical machine
  translation}.
\newblock In \emph{Proc. of EMNLP}.

\bibitem[{Devlin et~al.(2019)Devlin, Chang, Lee, and Toutanova}]{Devlin:2018}
Jacob Devlin, Ming{-}Wei Chang, Kenton Lee, and Kristina Toutanova. 2019.
\newblock {BERT:} pre-training of deep bidirectional transformers for language
  understanding.
\newblock In \emph{Proc. of NAACL}.

\bibitem[{Foerster et~al.(2017)Foerster, Gilmer, Chorowski, Sohl{-}Dickstein,
  and Sussillo}]{Foerster:2017}
Jakob~N. Foerster, Justin Gilmer, Jan Chorowski, Jascha Sohl{-}Dickstein, and
  David Sussillo. 2017.
\newblock Intelligible language modeling with input switched affine networks.
\newblock In \emph{Proc. of ICML}.

\bibitem[{Frankle and Carbin(2019)}]{Frankle:2019}
Jonathan Frankle and Michael Carbin. 2019.
\newblock The lottery ticket hypothesis: Finding sparse, trainable neural
  networks.
\newblock In \emph{Proc. of ICLR}.

\bibitem[{Gordon et~al.(2018)Gordon, Eban, Nachum, Chen, Wu, Yang, and
  Choi}]{morphnet}
Ariel Gordon, Elad Eban, Ofir Nachum, Bo~Chen, Hao Wu, Tien-Ju Yang, and Edward
  Choi. 2018.
\newblock {MorphNet}: Fast \& simple resource-constrained structure learning of
  deep networks.
\newblock In \emph{Proc. of CVPR}.

\bibitem[{de~la Higuera(2010)}]{higuera2010grammatical}
Colin de~la Higuera. 2010.
\newblock \emph{Grammatical Inference: Learning Automata and Grammars}.
\newblock Cambridge University Press.

\bibitem[{Hochreiter and Schmidhuber(1997)}]{Hochreiter:1997}
Sepp Hochreiter and J{\"u}rgen Schmidhuber. 1997.
\newblock \href {https://doi.org/10.1162/neco.1997.9.8.1735} {Long short-term
  memory}.
\newblock \emph{Neural computation}, 9(8).

\bibitem[{Johnson and Zhang(2015)}]{johnson2015effective}
Rie Johnson and Tong Zhang. 2015.
\newblock \href {https://doi.org/10.3115/v1/N15-1011} {Effective use of word
  order for text categorization with convolutional neural networks}.
\newblock In \emph{Proc. of NAACL}.

\bibitem[{Jozefowicz et~al.(2015)Jozefowicz, Zaremba, and
  Sutskever}]{Jozefowicz:2015}
Rafal Jozefowicz, Wojciech Zaremba, and Ilya Sutskever. 2015.
\newblock \href {http://www.jmlr.org/proceedings/papers/v37/jozefowicz15.pdf}
  {An empirical exploration of recurrent network architectures}.
\newblock In \emph{Proc. of ICML}.

\bibitem[{Kingma and Ba(2015)}]{Kingma:2014}
Diederik Kingma and Jimmy Ba. 2015.
\newblock Adam: A method for stochastic optimization.
\newblock In \emph{Proc. of ICLR}.

\bibitem[{LeCun et~al.(1990)LeCun, Denker, and Solla}]{brain_damage}
Yann LeCun, John~S. Denker, and Sara~A. Solla. 1990.
\newblock Optimal brain damage.
\newblock In \emph{NeurIPS}.

\bibitem[{Lee et~al.(2019)Lee, Ajanthan, and Torr}]{SNIP}
Namhoon Lee, Thalaiyasingam Ajanthan, and Philip~HS Torr. 2019.
\newblock {SNIP}: Single-shot network pruning based on connection sensitivity.
\newblock In \emph{Proc. of ICLR}.

\bibitem[{Lei et~al.(2017)Lei, Zhang, and Artzi}]{Lei:2017}
Tao Lei, Yu~Zhang, and Yoav Artzi. 2017.
\newblock Training {RNNs} as fast as {CNNs}.
\newblock {arXiv}:1709.02755.

\bibitem[{Liu et~al.(2019)Liu, Simonyan, and Yang}]{DARTS}
Hanxiao Liu, Karen Simonyan, and Yiming Yang. 2019.
\newblock {DARTS}: Differentiable architecture search.
\newblock In \emph{Proc. of ICLR}.

\bibitem[{Livni et~al.(2014)Livni, Shalev-Shwartz, and Shamir}]{Livni:2014}
Roi Livni, Shai Shalev-Shwartz, and Ohad Shamir. 2014.
\newblock On the computational efficiency of training neural networks.
\newblock In \emph{NeurIPS}.

\bibitem[{Martins et~al.(2011)Martins, Smith, Figueiredo, and
  Aguiar}]{Martins:2011}
Andr\'{e} F.~T. Martins, Noah~A. Smith, Mario Figueiredo, and Pedro Aguiar.
  2011.
\newblock \href {http://aclweb.org/anthology/D11-1139} {Structured sparsity in
  structured prediction}.
\newblock In \emph{Proc. of EMNLP}.

\bibitem[{Oncina et~al.(1993)Oncina, Garc{\'i}a, and
  Vidal}]{oncina1993learning}
Jos{\'e} Oncina, Pedro Garc{\'i}a, and Enrique Vidal. 1993.
\newblock Learning subsequential transducers for pattern recognition
  interpretation tasks.
\newblock \emph{IEEE Trans. Pattern Anal. Mach. Intell.}, 15:448--458.

\bibitem[{Parikh and Boyd(2013)}]{parikhboyd2013}
Neal Parikh and Stephen~P. Boyd. 2013.
\newblock \emph{Proximal Algorithms}.
\newblock Foundations and Trends in Optimization.

\bibitem[{Peng et~al.(2018)Peng, Schwartz, Thomson, and
  Smith}]{rational-recurrences}
Hao Peng, Roy Schwartz, Sam Thomson, and Noah~A. Smith. 2018.
\newblock \href {http://aclweb.org/anthology/D18-1152} {Rational recurrences}.
\newblock In \emph{Proc. of EMNLP}.

\bibitem[{Pennington et~al.(2014)Pennington, Socher, and
  Manning}]{Pennington:2014}
Jeffrey Pennington, Richard Socher, and Christopher Manning. 2014.
\newblock \href {http://www.aclweb.org/anthology/D14-1162} {Glo{V}e: Global
  vectors for word representation}.
\newblock In \emph{Proc. of EMNLP}.

\bibitem[{Radford et~al.(2019)Radford, Wu, Child, Luan, Amodei, and
  Sutskever}]{Radford:2019}
Alec Radford, Jeffrey Wu, Rewon Child, David Luan, Dario Amodei, and Ilya
  Sutskever. 2019.
\newblock Language models are unsupervised multitask learners.

\bibitem[{Reddi et~al.(2016)Reddi, Sra, Póczos, and Smola}]{reddiProximal}
Sashank~J. Reddi, Suvrit Sra, Barnabás Póczos, and Alexander~J. Smola. 2016.
\newblock Proximal stochastic methods for nonsmooth nonconvex finite-sum
  optimization.
\newblock In \emph{NeurIPS}.

\bibitem[{Ron et~al.(1994)Ron, Singer, and Tishby}]{ron1994learning}
Dana Ron, Yoram Singer, and Naftali Tishby. 1994.
\newblock Learning probabilistic automata with variable memory length.
\newblock In \emph{Proc. of COLT}.

\bibitem[{Scardapane et~al.(2017)Scardapane, Comminiello, Hussain, and
  Uncini}]{group_lasso_nn}
Simone Scardapane, Danilo Comminiello, Amir Hussain, and Aurelio Uncini. 2017.
\newblock Group sparse regularization for deep neural networks.
\newblock \emph{Neurocomputing}, 241.

\bibitem[{Schwartz et~al.(2019)Schwartz, Dodge, Smith, and
  Etzioni}]{Schwartz:2019}
Roy Schwartz, Jesse Dodge, Noah~A. Smith, and Oren Etzioni. 2019.
\newblock \href {https://arxiv.org/abs/1907.10597} {Green {AI}}.
\newblock {arXiv}:1907.10597.

\bibitem[{Schwartz et~al.(2018)Schwartz, Thomson, and Smith}]{Schwartz:2018}
Roy Schwartz, Sam Thomson, and Noah~A. Smith. 2018.
\newblock \href {http://aclweb.org/anthology/P18-1028} {{SoPa}: Bridging
  {CNNs}, {RNNs}, and weighted finite-state machines}.
\newblock In \emph{Proc. of ACL}.

\bibitem[{Strubell et~al.(2019)Strubell, Ganesh, and McCallum}]{Strubell:2019}
Emma Strubell, Ananya Ganesh, and Andrew McCallum. 2019.
\newblock Energy and policy considerations for deep learning in {NLP}.
\newblock In \emph{Proc. of ACL}.

\bibitem[{Wen et~al.(2016)Wen, Wu, Wang, Chen, and Li}]{wen2016learning}
Wei Wen, Chunpeng Wu, Yandan Wang, Yiran Chen, and Hai Li. 2016.
\newblock Learning structured sparsity in deep neural networks.
\newblock In \emph{NeurIPS}.

\bibitem[{Yuan and Lin(2006)}]{group_lasso}
Ming Yuan and Yi~Lin. 2006.
\newblock Model selection and estimation in regression with grouped variables.
\newblock \emph{Journal of the Royal Statistical Society: Series B (Statistical
  Methodology)}, 68(1).

\bibitem[{Zoph and Le(2017)}]{NAS}
Barret Zoph and Quoc~V Le. 2017.
\newblock Neural architecture search with reinforcement learning.
\newblock In \emph{Proc of ICLR}.

\end{thebibliography}
\bibliographystyle{acl_natbib}

\clearpage
\appendix
\section{Experiment Details}\label{app:experimental_details}

\paragraph{Dataset statistics}
 \tabref{amazon_sizes} shows the sizes of the datasets experimented with.
 
\begin{table}[h]
  \centering
  \small
  \begin{tabular}{@{}l r r r@{}}
\toprule
& {\bf Training} & {\bf Dev.} & {\bf Test}\\
\midrule
{\bf kitchen} & 3,298 & 822 & 4,118\\
{\bf dvd} & 14,066 & 3,514 & 17,578\\
{\bf books} & 20,000 & 5,000 & 25,000 \\
{\bf original\_mix} & 20,000 & 5,000 & 25,000 \\
\bottomrule
\end{tabular}
   \caption{Text classification dataset sizes. 
   	Each dataset follows the same training/dev./test split ratio as the original mix.}
  \label{tab:amazon_sizes}
\end{table}

\paragraph{Preprocessing}

As preprocessing for the data for each individual category, we
tokenize using the NLTK word tokenizer. We removed reviews with text shorter than 5 tokens.

We binarize the review score using the standard procedure, assigning 1- and 2-star reviews as negative, and 4- and 5-star reviews as positive (discarding 3-star reviews). 
Then, if there were more than 25,000 negative reviews, we downsample to 25,000 (otherwise we keep them all), and then downsample the positive reviews to be the same number as negative, to have a balanced dataset. We match the train, development, and test set proportions of 4:1:5 from the original mixture. 

We generate the BERT embeddings using the sum of the last four hidden layers of the large uncased BERT model, so our embedding size is 1024. {\com{Finetuning BERT was too computationally expensive, so we just kept the embeddings. }Summing the last four layers was the best performing approach in the ablation of \citet{Devlin:2018} that had fewer than 4096 embedding size (which was too large to fit in memory).}
We embed each sentence individually (there can be multiple sentences within one example).

\paragraph{Implementation details}
For GloVe, we train rational models with 24 5-state WFSAs, each corresponding to a 4-gram soft-pattern (\figref{4gram_wfsa}).
For BERT, we train models with 12 WFSAs.\footnote{The BERT embedding dimension is significantly larger than GloVe (1024 compared to 300), so we used a smaller number of WFSAs. As our results show, the BERT models still substantially outperform the GloVe ones.}  

\paragraph{Experiments}
For each model (regularized or baseline), we run random search to select our  hyperparameters (evaluating 20 uniformly sampled hyperparameter configurations). 
For the hyperparameter configuration that leads to the best development result, we train the model again 5 times with different random seeds, and report the mean and standard deviation of the models' test performance.

\paragraph{Parameters}

The models are trained with Adam \cite{Kingma:2014}. During training with group lasso we turn off the learning rate schedule (so the learning rate stays fixed), similarly to \citet{morphnet}. This leads to improved stability in the learned structure for a given hyper-parameter assignment.

Following \citet{rational-recurrences} we sample 20 hyperparameters uniformly, for which we train and evaluate our models.
Hyperparameter ranges are presented in Table \ref{tab:hparam_ranges}. For the BERT experiments, we reduced both the upper and lower bound on the learning rate by two orders of magnitude.

\paragraph{Regularization strength search}
We searched for model structures that were regularized down to close to 20, 40, 60, or 80 transitions (10, 20, 30, and 40 for BERT experiments). 
For a particular goal size, we uniformly sample 20 hyperparameter assignments from the ranges in Table~\ref{tab:hparam_ranges}, then sorted the samples by increasing learning rate. 
For each hyperparameter assignment, we trained a model with the current regularization strength. 
If the resulting learned structure was too large (small), we doubled (halved) the regularization strength, repeating until we were within 10 transitions of our goal (5 for BERT experiments).\footnote{If the regularization strength became larger than $10^2$ or smaller than $10^{-9}$, we threw out the hyperparameter assignment and resampled (this happened when, e.g., the learning rate was too small for any of the weights to actually make it to zero).}
Finally, we finetuned the appropriately-sized learned structure by continuing training without the regularizer, and computed the result on the development set.
For the best model on the development set, we retrained (first with the regularizer to learn a structure, then finetuned) five times, and plot the mean and variance of the test accuracy and learned structure size.

\section{Visualization}\label{app:visualization}

\begin{table}[t]
\small
\newcommand{\sloop}[1]{{\color{blue}{#1$_{\textit{SL}}$}} }
\setlength{\tabcolsep}{4.7pt}
\scriptsize
\begin{tabularx}{\linewidth}{c|c|lll}
 \toprule
\multicolumn{2}{c}{} &  {\bf transition$_1$} &     {\bf transition$_2$}        &       {\bf transition$_3$}  \\
  \toprule

\multirow{10}{*}{\vspace{0.5cm} Patt.~1} &\multirow{5}{*}{\vspace{0.25cm} \highest} & are & perfect & \sloop{\dots} [CLS]  \\
& &definitely & recommend & \sloop{\dots} [CLS]  \\
& &excellent & product & \sloop{\dots} [CLS]  \\
& &highly & recommend & \sloop{\dots} [CLS]  \\
\cline{2-5}

&\multirow{5}{*}{\vspace{0.25cm} \bottom} & not & \sloop{\dots} [SEP] & \sloop{\dots} [CLS]  \\
& &very & disappointing & \sloop{!} \sloop{[SEP]} [CLS]  \\
& &was & defective & \sloop{\dots} had  \\
& &would & not & \sloop{\dots} [CLS]  \\
\midrule

\multirow{10}{*}{\vspace{0.5cm} Patt.~2} &\multirow{5}{*}{\vspace{0.25cm} \highest} & [CLS] & mine & broke  \\
& &[CLS] & it & \sloop{\dots} heat  \\
& &[CLS] & thus & it  \\
& &[CLS] & \sloop{it} does & \sloop{it} heat  \\
\cline{2-5}

&\multirow{5}{*}{\vspace{0.25cm} \bottom} & [CLS] & perfect & \sloop{\dots} cold  \\
& &[CLS] & sturdy & \sloop{\dots} cooks  \\
& &[CLS] & evenly & \sloop{,} \sloop{withstand} heat  \\
& &[CLS] & it & is  \\
\midrule

\multirow{10}{*}{\vspace{0.5cm} Patt.~3} &\multirow{5}{*}{\vspace{0.25cm} \highest} & ` & pops & \sloop{'} \sloop{'} escape  \\
& &` & gave & out  \\
& &that & had & escaped  \\
& &` & non & -  \\
\cline{2-5}

&\multirow{5}{*}{\vspace{0.25cm} \bottom} & simply & does & not  \\
& &[CLS] & useless & \sloop{equipment} !  \\
& &unit & would & not  \\
& &[CLS] & poor & \sloop{to} no  \\

\midrule

\multirow{10}{*}{\vspace{0.5cm} Patt.~4} &\multirow{5}{*}{\vspace{0.25cm} \highest} & [CLS] & after &  \\
& &[CLS] & our &  \\
& &mysteriously & jammed &  \\
& &mysteriously & jammed &  \\
\cline{2-5}

&\multirow{5}{*}{\vspace{0.25cm} \bottom} & [CLS] & i &  \\
& &[CLS] & i &  \\
& &[CLS] & i &  \\
& &[CLS] & we &  \\
\bottomrule
\end{tabularx}
\caption{\label{tab:vizb}
Visualization of a sparse rational RNN containing 4 WFSAs only, trained on {\bf kitchen} using BERT.
}

\end{table}

\tabref{vizb} shows the same visualization shown in \secref{visualization} for another sparse rational RNN containing only four WFSAs and 11 main-path transitions, trained with BERT embeddings on {\bf kitchen}.
It also shows a few clear patterns (e.g., Patt.~2).
Interpretation here is more challenging though, as contextual
embeddings make every token embedding depend on the entire context\resolved{\nascomment{this is a really nice point.  do we
  have more to say about it?  e.g., implications for interpretability
  with contextual embeddings!?}}.\footnote{Indeed, contextual embeddings raise problems for interpretation methods that work by targeting individual words, e.g., attention \cite{Bahdanau:2015}, as these embeddings also depend on other words. Interpretation methods for contextual embeddings are an exciting direction for future work.
}
A particular example of this is the excessive use of the start token
([CLS]), whose contextual embedding has been shown to capture the
sentiment information at the sentence level
\cite{Devlin:2018}.

\paragraph{Regularization strength recommendation}
If a practitioner wishes to learn a single small model, we recommend
they start with $\lambda$ such that the loss $\mathcal{L}(\mathbf{w})$
and the regularization term are equal at initialization (before training). \resolved{\nascomment{at convergence?}\jdcomment{actually, at initialization.}}
We found that having equal contribution led to eliminating approximately half of the states, though this varies with data set size, learning rate, and gradient clipping, among other variables.

\begin{table}[!t]
  \centering
  \small
  \begin{tabularx}{\linewidth}{@{}X X@{}}
\toprule
\bf{Type} & \bf{Range}\\
\midrule
Learning rate & [$7*10^{-3}$, $0.5$] \\
Vertical dropout & [0, 0.5] \\
Recurrent dropout & [0, 0.5] \\
Embedding dropout & [0, 0.5] \\
$\ell_2$ regularization & [0, 0.5] \\
Weight decay & [$10^{-5}$, $10^{-7}$]\\
\bottomrule
\end{tabularx}
   \caption{Hyperparameter ranges considered in our experiments.}
  \label{tab:hparam_ranges}
\end{table}


\end{document}